\documentclass{article}
\usepackage{amssymb}
\usepackage{graphicx}
\usepackage{amsmath}
\usepackage{booktabs}
\usepackage{array}
\usepackage{amsmath}
\usepackage{mathrsfs}
\usepackage[final]{corl_2025} 

\title{NeRF-Aug: Data Augmentation for Robotics with Neural Radiance Fields}

\author{
  Eric Zhu, Mara Levy, Matthew Gwilliam, Abhinav Shrivastava\\
  Department of Computer Science\\
  University of Maryland College Park\\
  \texttt{\{ezhu2958,mlevy,mgwillia,abhinav2\}@umd.edu}
}

\begin{document}
\maketitle

\begin{abstract}
    Training a policy that can generalize to unknown objects is a long standing challenge within the field of robotics. The performance of a policy often drops significantly in situations where an object in the scene was not seen during training. To solve this problem, we present NeRF-Aug, a novel method that is capable of teaching a policy to interact with objects that are not present in the dataset. This approach differs from existing approaches by leveraging the speed, photorealism, and 3D consistency of a neural radiance field for augmentation. NeRF-Aug both creates more photorealistic data and runs 63\% faster than existing methods. We demonstrate the effectiveness of our method on 5 tasks with 9 novel objects that are not present in the expert demonstrations. We achieve an average performance boost of 55.6\% when comparing our method to the next best method. You can see video results at \url{https://nerf-aug.github.io}.
\end{abstract}


\section{Introduction}
Humans have an innate ability to interact with objects they have never encountered before.
For instance, a person can intuitively approach an unknown object, pick it up, and interact with it. 
This is in stark contrast to existing robotic systems. 
Even the slightest differences in shape or color from the objects seen during training can prevent the robot from achieving success. This challenge of generalization to out-of-distribution samples is a fundamental issue in machine learning and robotics.

Many prior works have explored methods to develop policies for robots that generalize to different objects.
A straightforward approach is to simply collect demonstrations involving the novel object. However, this method has significant drawbacks because creating expert demonstrations is time-consuming and expensive as it requires a human to consciously control the robot's movements. Therefore, collecting such human demonstrations is infeasible at scale. 

Another approach is to use image editing tools, e.g., the latest diffusion-based image editing~\cite{bharadhwaj2023roboagentgeneralizationefficiencyrobot, chen2023genaugretargetingbehaviorsunseen, mandi2023cactiframeworkscalablemultitask, Yu2023ScalingRL, Kapelyukh_2023}. While these models can effectively edit images to insert new objects, they are often slow and struggle to render the exact object that will be encountered by the robot. This inaccuracy means the current object remains out of the domain of the training set which often causes these models to fail.
Alternatively, some pipelines use depth images for object manipulation~\cite{Zhu20216DoFCG, 8793820}.
Unfortunately, depth images in the real world suffer from noise and incompleteness~\cite{8793820}. 
This issue is exacerbated when using mounted gripper cameras, which amplify noise as they get closer to the object. Moreover, even though depth images disregard texture and color, small geometric differences between the original (training) and novel objects can still result in confusion.

\begin{figure}
    \centering
    \includegraphics[width=0.45\linewidth, trim=0 0 0 100]{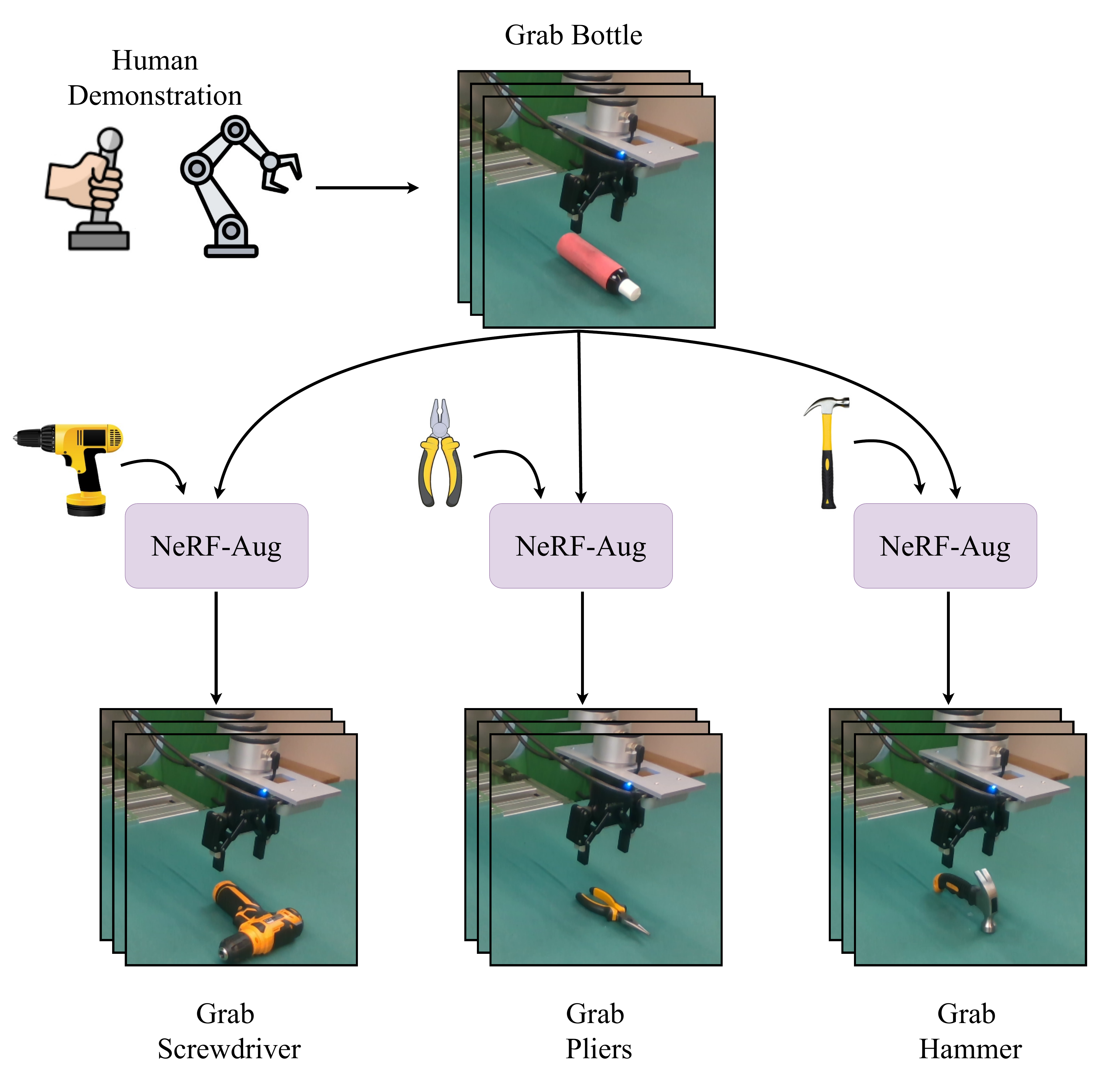}
    \vspace{-0.2in}
    \caption{When a human provides expert demonstrations for training a behavior cloning model, the model is effective for the object in the demonstration, but will fail for novel objects. We propose NeRF-Aug, where we automatically learn NeRFs for the novel objects and inpaint them in the expert data. With this photorealistic synthetic data, the robot can learn to interact with novel objects.}
    \label{fig:teaser}
    \vspace{-0.2in}
\end{figure}

In this work, we propose NeRF-Aug, a lightweight framework that streamlines and automates data collection for a wide range of novel objects. Our goal is to generate data samples for different tasks using novel objects without collecting more human demonstrations. To enable this, we follow the image editing paradigm, but instead of relying on slow generation frameworks, we propose using a 3D model of a novel object with a Neural Radiance Field (NeRF)~\cite{mildenhall2020nerfrepresentingscenesneural} representation. We augment the training data for the robot's policy using this edited scene.
Our framework uses existing demonstrations of a different object and generates NeRF-Augmented (NeRF-Aug) synthetic data (Fig.~\ref{fig:overview}) that can be used in imitation learning policies.

We demonstrate that the synthetic data generated by the NeRF-Aug framework is almost indistinguishable from real-world data. Moreover, we show that compared to existing diffusion-based image editing techniques, our method runs significantly faster, creates photorealistic images, and can consistently render objects at a wide degree of viewpoints.

We test our method on a variety of real world tasks, and achieve a 55.6\% increase in success rate over generation based approaches while rendering the synthetic data 63\% faster than the baselines.

To summarize, our contributions are as follows:
\begin{itemize}
    \item We propose a fast and photorealistic image editing framework to generate synthetic data that can be used in robot policy learning to generalize to novel objects.
    \item We learn a NeRF of a novel object by using multi-view images of the object captured using a robot arm.
    \item We edit existing expert video demos by removing the training object via in-painting and blend the NeRF render of a novel object into the inpainted image to create synthetic data.
    \item We demonstrate effective generalization on five diverse tasks using the generated synthetic dataset for training.
\end{itemize}

\section{Related Works}
\subsection{Imitation Leaning}
Imitation learning is a common approach to creating robotic policies. Several approaches have been proposed that allow a policy to learn from existing successful trajectories.~\cite{bharadhwaj2023roboagentgeneralizationefficiencyrobot, brohan2023rt1roboticstransformerrealworld, haldar2024bakuefficienttransformermultitask, torabi2018behavioralcloningobservation} introduce various methods that take observations as input and predict actions. Of these,~\cite{bharadhwaj2023roboagentgeneralizationefficiencyrobot, haldar2024bakuefficienttransformermultitask} are multi-input models that input values from different types of data at the same time.~\cite{zhang2024diffusionmeetsdaggersupercharging, kelly2019hgdaggerinteractiveimitationlearning, hoque2022fleetdaggerinteractiverobotfleet, sun2024megadaggerimitationlearningmultiple} go a step further and use an expert to actively correct the policy when as it executes trajectories.~\cite{swamy2024inversereinforcementlearningreinforcement, ho2016generativeadversarialimitationlearning, reddy2019sqilimitationlearningreinforcement, fu2018learningrobustrewardsadversarial} take initial demonstrations and learn a reward function to train a reinforcement learning policy.

\subsection{Neural Radiance Fields (NeRFs)}
In recent years, there has been a surge in research on NeRFs.~\cite{wang2023f2nerffastneuralradiance, Reiser2021KiloNeRFSU, Hedman2021BakingNR, RivasManzaneque2023NeRFLightFA, Lin2021EfficientNR} are all NeRF approaches that are capable of rendering high resolution images in real time. Other methods focus on the robustness of NeRF models against real-world constraints, such as limited number of images~\cite{gu2023nerfdiffsingleimageviewsynthesis}, camera pose noise~\cite{pearl2022nannoiseawarenerfsburstdenoising}, and glare~\cite{guo2022nerfrenneuralradiancefields}. Finally, other novel-view synthesis methods~\cite{Yu2021PlenoxelsRF, Yu2021PlenOctreesFR, Sun2021DirectVG, kerbl20233dgaussiansplattingrealtime} remove the traditional mlp used to render images with other trainable structures.

Within the context of robotics, NeRFs have been used for robot navigation by creating a 3D map of the environment ~\cite{byravan2022nerf2realsim2realtransfervisionguided}.~\cite{shim2023snerlsemanticawareneuralradiance, driess2022reinforcementlearningneuralradiance} combine NeRFs and reinforcement learning. ~\cite{lee2022uncertaintyguidedpolicyactive, chen2023perceivingunseen3dobjects} explored using robot arms for creating higher quality NeRF models of objects. Conversely, several works have explored using NeRF models for robot decision making, including~\cite{kerr2022evonerf, ichnowski2021dexnerfusingneuralradiance} which creates accurate depth maps for transparent images using NeRF models, and~\cite{kapelyukh2024dream2realzeroshot3dobject} which creates an object level NeRF and imagines a scene with the object in a different place and queries a vision-language model (VLM) for feedback. Finally,~\cite{zhou2023nerfpalmhandcorrective} adds noisy viewpoints along a demonstration trajectory and take corrective actions to make a more robust behavior cloning model.

\subsection{Synthetic Data in Robotics}
Creating synthetic data for robot learning is a popular paradigm for training data hungry machine learning models.~\cite{jiang2022dittobuildingdigitaltwins, torne2024reconcilingrealitysimulationrealtosimtoreal} champion creating digital twins of the current robot environments and running the robot in these simulations.~\cite{hsu2023dittohousebuildingarticulation} go a step further and automate the exploration of real-world environments to create the high-fidelity digital twin.~\cite{byravan2022nerf2realsim2realtransfervisionguided} combine a simulator with a NeRF model to create a photorealistic policy that could bridge the sim-to-real gap.\looseness=-1

Another direction for using synthetic data is using diffusion models in robotics.~\cite{bharadhwaj2023roboagentgeneralizationefficiencyrobot, chen2023genaugretargetingbehaviorsunseen, mandi2023cactiframeworkscalablemultitask, Yu2023ScalingRL, Kapelyukh_2023} use diffusion models to randomize the texture of objects and swap objects in a scene.~\cite{du2023learninguniversalpoliciestextguided, huang2024ardupactiveregionvideo} use a video diffusion model to create synthetic videos of a robot performing a task and then let the robot recreate these videos. ~\cite{black2023zeroshotroboticmanipulationpretrained} use a text-to-image diffusion model to synthesize a goal image of objects in a desired location for a goal conditioned reinforcement learning model to subsequently rearrange.~\cite{zhang2024diffusionmeetsdaggersupercharging} uses diffusion to model expert online corrections. Finally \cite{chen2024roviaugrobotviewpointaugmentation} uses diffusion to switch the model of robot arm performing an action in a demonstration.
\section{Preliminaries}
\noindent\textbf{Imitation Learning}. In imitation learning, we assume a premade list of $N$ expert trajectories $D = \left\{\tau_i\right\}_{i=1}^N$ where each trajectory consists of state-action pairs $\tau_i = \{(s_k, a_k)\}_{k=1}^K$. In visual imitation learning, we assume access to $n$ cameras where each contributes an image to the state. Thus, our state can be denoted as $s_k = (I_1, I_2, ..., I_n)$ where $I_k$ is an image. In vanilla behavior cloning, the policy $\pi$ is trained offline and learns a mapping between states and actions from our expert dataset by minimizing the loss $\mathcal{L}(\theta) = \mathbb{E}_{(s,a) \sim D}\left[\ell\left(\pi(s), a\right)\right]$ for a given distance metric $\ell$.

\noindent\textbf{Neural Radiance Fields}. NeRF is a method used for rendering novel views of a scene or object. Formally, we assume a dataset of images and corresponding camera poses $D = \{(I_k, T_k)\}_{k=1}^K$ where $T_k \in \text{SE}(3)$ and $I_k \in \mathbb{R}^{w \times h \times 3}$. Neural radiance fields are able to render the scene at a novel pose $T$ through volumetric rendering. The result is a photorealistic image of a given scene by querying the NeRF model with a requested camera pose.

\begin{figure*}
    \centering
    \vspace{-0.1in}
    \includegraphics[width=0.9\linewidth]{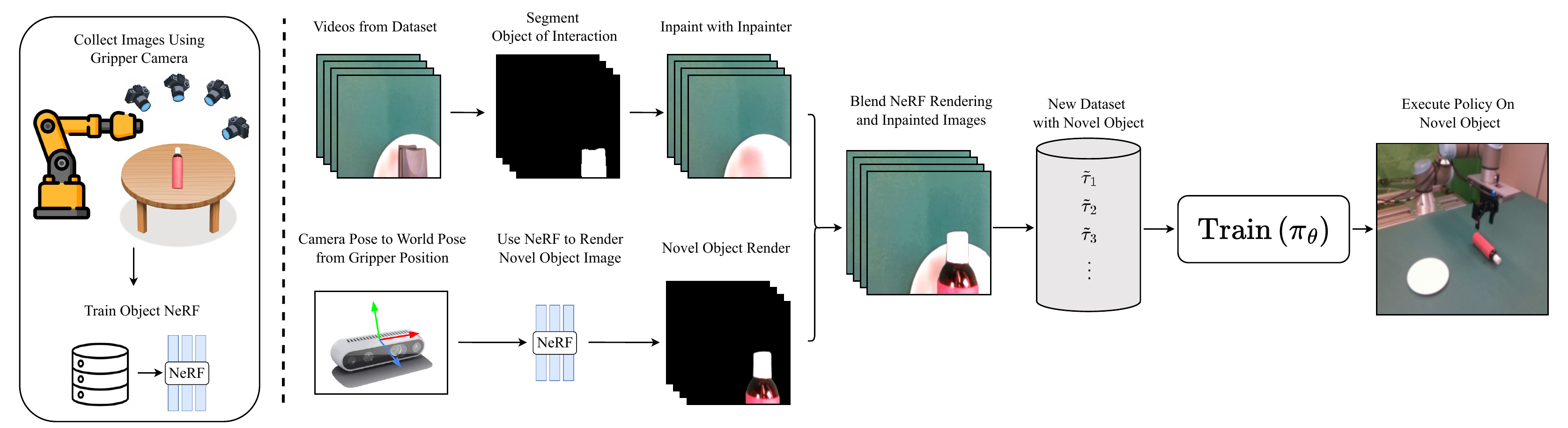}
    \vspace{-0.1in}
    \caption{An illustration of our pipeline from beginning to end. We first train an object-level NeRF of a novel object (\textbf{left}). We then simultaneously erase the object in question with an inpainter (\textbf{top}) and leverage NeRF to render images of a new object in the same position as the original object (\textbf{bottom}). We use the final synthetic dataset to train a new policy for the robot (\textbf{right}).}
    \label{fig:overview}
    \vspace{-0.15in}
\end{figure*}
\section{Our Approach: NeRF-Aug}
The first step in our approach is capturing images of a novel object from multiple viewpoints and training a corresponding NeRF model. Next, we use the robot arm's gripper position to calculate the camera pose at each image in our trajectory. Using these camera poses we query the NeRF model to render an image of the novel object at the same position and orientation as our original object in the training example. Then, we combine the NeRF rendering and the original image to create a new synthetic image of the robot handling the novel object. Finally, we train our policy on this new dataset and evaluate the original task on the novel object. An illustrative overview of our proposed NeRF-Aug approach is shown in Fig.~\ref{fig:overview}. 

\subsection{Creating a NeRF model of the Novel Object} Given a novel object for which there exists no training data, we train a NeRF model for this object. Using a gripper mounted camera, we are able to collect a dataset of image-pose pairs $\left\{(I_k, T_k)\right\}_{k=1}^M$ by moving the gripper to various viewpoints both close and far away from the novel object. We use these images to train a NeRF model for the object. In our experiments, we used NerfStudio's~\cite{Tancik_2023} default Nerfacto model, which refines the camera parameters of all images to denoise measurements.

\subsection{Calculating Camera Pose Relative to Object}
\label{subsec:calculate_pose}
Our method relies on accurately rendering a new object in the same position as the original training object at each frame. For simplicity we use a set position in this project, but the original object position can easily be calculated from the gripper position at the time the object is first grasped.

Next, NeRF-Aug finds the camera position relative to the object. Using the gripper position with respect to the world given by $T_\text{gripper}(t) \in \text{SE}(3)$, we are able to find the camera-to-world matrix of the gripper camera. We multiply the gripper position by the offset of the camera from the gripper
\begin{equation}
    T_\text{camera-to-world}(t) = T_\text{gripper}(t) \cdot T_\text{camera-offset}
\end{equation}
The relative position of the camera with respect to the object coordinate system is denoted by \begin{equation}
        T_\text{camera-to-object}(t) = T_\text{object-to-world}^{-1} (t) \cdot T_\text{camera-to-world}(t)
\end{equation}
If the object is grasped, the relative position of the camera and the object stays the same, denoted by
\begin{equation}
        T_\text{camera-to-object}(t) = T_\text{camera-to-object}(t_\text{grasp})\text{ for $t \geq t_\text{grasp}$}.
\end{equation}
We also add a small amount of noise at each timestep into the camera-to-object matrix before rendering in the NeRF model. This slightly shifts the object from its original position to simulate a larger degree of grasp positions. 

\subsection{NeRF Renderings}
\label{subsec:nerf_renderings}
\indent For each frame in the trajectory, we produce a rendering of our novel object at the same position and orientation as the original object by plugging our camera-to-object matrix into the NeRF model, $I_{\text{NeRF}} = \text{NeRF\_Render}\left(T_{\text{camera-to-object}}(t)\right)$. Because we are only interested in the novel object rendered from the NeRF and do not need the background of the NeRF renderings, we find a mask for the object denoted as $M_\text{NeRF}$. 

\begin{figure*}
    \centering
    \includegraphics[width=0.6\linewidth, trim=0 50 0 50,
                 clip]{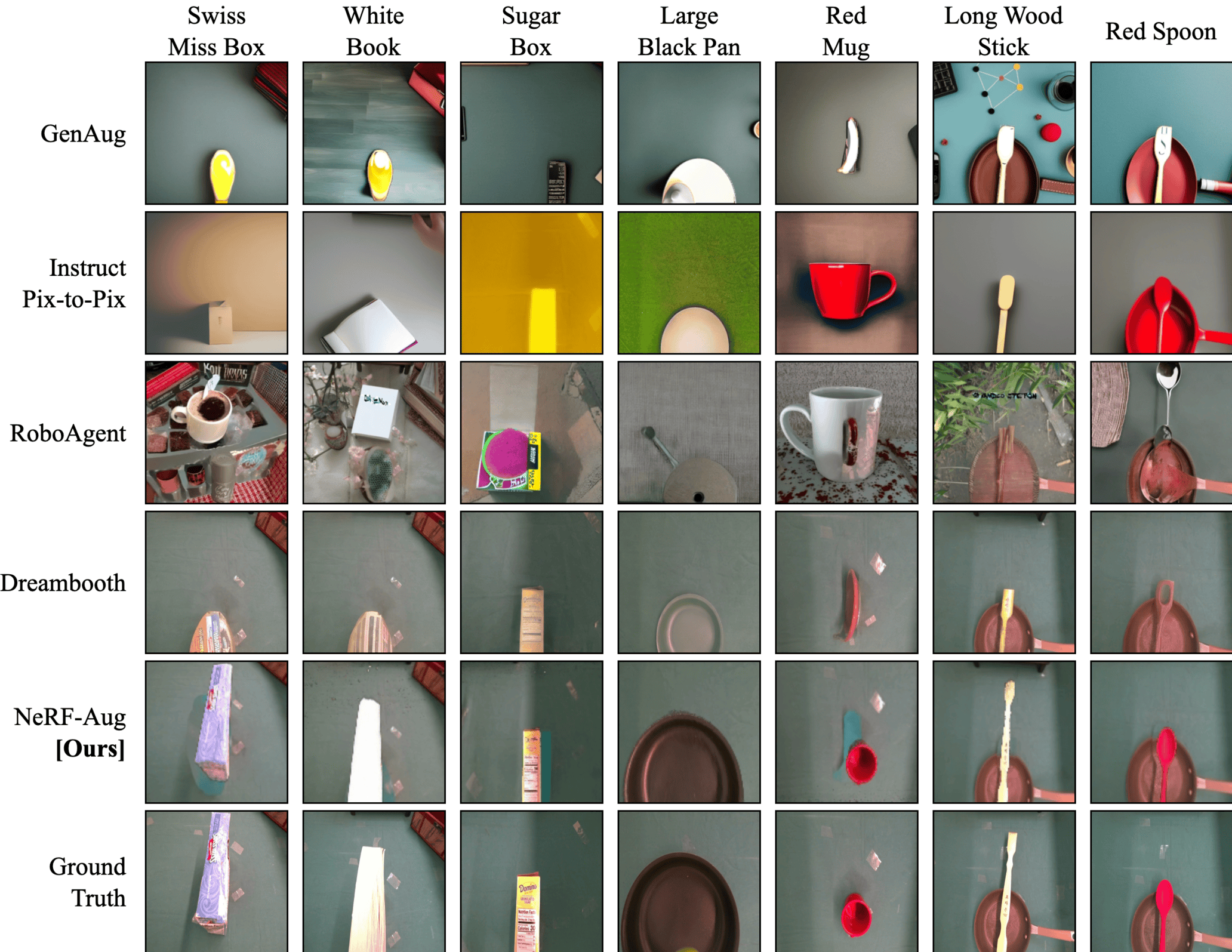}
    \vspace{-0.1in}
    \caption{Inpainting results when replacing the original object with various objects in our training set. Instruct P2P, GenAug, and RoboAgent create images that show a new object, but do not recreate our specific object with appropriate dimensions and texture, causing these methods to fail. Dreambooth has trouble with inpainting complex masks and does not recreate the right shade of colors.}
    \label{fig:results2}
    \vspace{-0.1in}
\end{figure*}

\subsection{Combining NeRF Renderings and Original Image}
\indent Once we have the NeRF rendering of the new object and the original image, we need to combine these two images. Our new object is potentially smaller than the original object, so we cannot simply copy over all the object pixels from the NeRF rendering onto our original image. If we were to do so, pixels from the original object could still remain in the new synthetic image. Thus, we choose to erase the original object using an inpainter. While any inpainter would work, we chose to inpaint using a NeRF model trained on the background of the scene. To do this, we use a pretrained Segment-Anything model~\cite{kirillov2023segment} to segment the original object, Cutie video object tracker~\cite{cheng2024puttingobjectvideoobject} to track the segmentation mask through the video frames, and a background NeRF model to switch the pixels in this mask with pixels from the background. 

This results in a frame where the original object is removed from the frame and only the background scene remains. From there, we blend the object pixels from the NeRF renderings onto the image where the object was removed. We combine $I_\text{NeRF}$ and $I_\text{no-object}$ to get a new image with the novel object at the same pose as the original object while still keeping the background of the original image. This is done by computing \begin{equation}
    I_\text{final} = I_\text{NeRF} \odot M_\text{NeRF} + I_\text{no-object} \odot \left(1 - M_\text{NeRF}\right).
\end{equation}
These frames and corresponding actions can then be trained using a behavior cloning model and run on a scenario involving the new object.
\begin{figure*}[!t]
    \vspace{-0.3in}
    \centering\includegraphics[width=0.66\textwidth]{figures/Results1_With_Dreambooth_compressed.jpg} 
    \caption{A comparison of data augmentation results for the ``Remove Object'' task. The top shows various techniques and how they compare to a original trajectory. Below that we show four methods that generate an expert synthetic trajectory for two novel objects (red spoon on the left, blue tongs on the right). GenAug is effective at swiping the texture of the object, but the color stays consistent with the original object. Instruct Pix-to-Pix is better at switching the color but also heavily changes the shading. RoboAgent adds a bunch of distractor objects that could confuse a behavior cloning model. Dreambooth inpaints an object at the same dimensions at the original object and in a different shade. Our method creates nearly identical trajectories to those collected in the real world.}
    \label{fig:results1}
    \vspace{-0.1in}
\end{figure*}
\section{Experiments}
We test our method on 5 real-world tasks: 1) \textbf{Remove Object}, 2) \textbf{Place in Chest}, 3) \textbf{Serve On Plate}, 4) \textbf{Cook with Pot}, and 5) \textbf{Lift up High}. 

\begin{figure}[t]
    \centering
    \includegraphics[width=.25\linewidth]{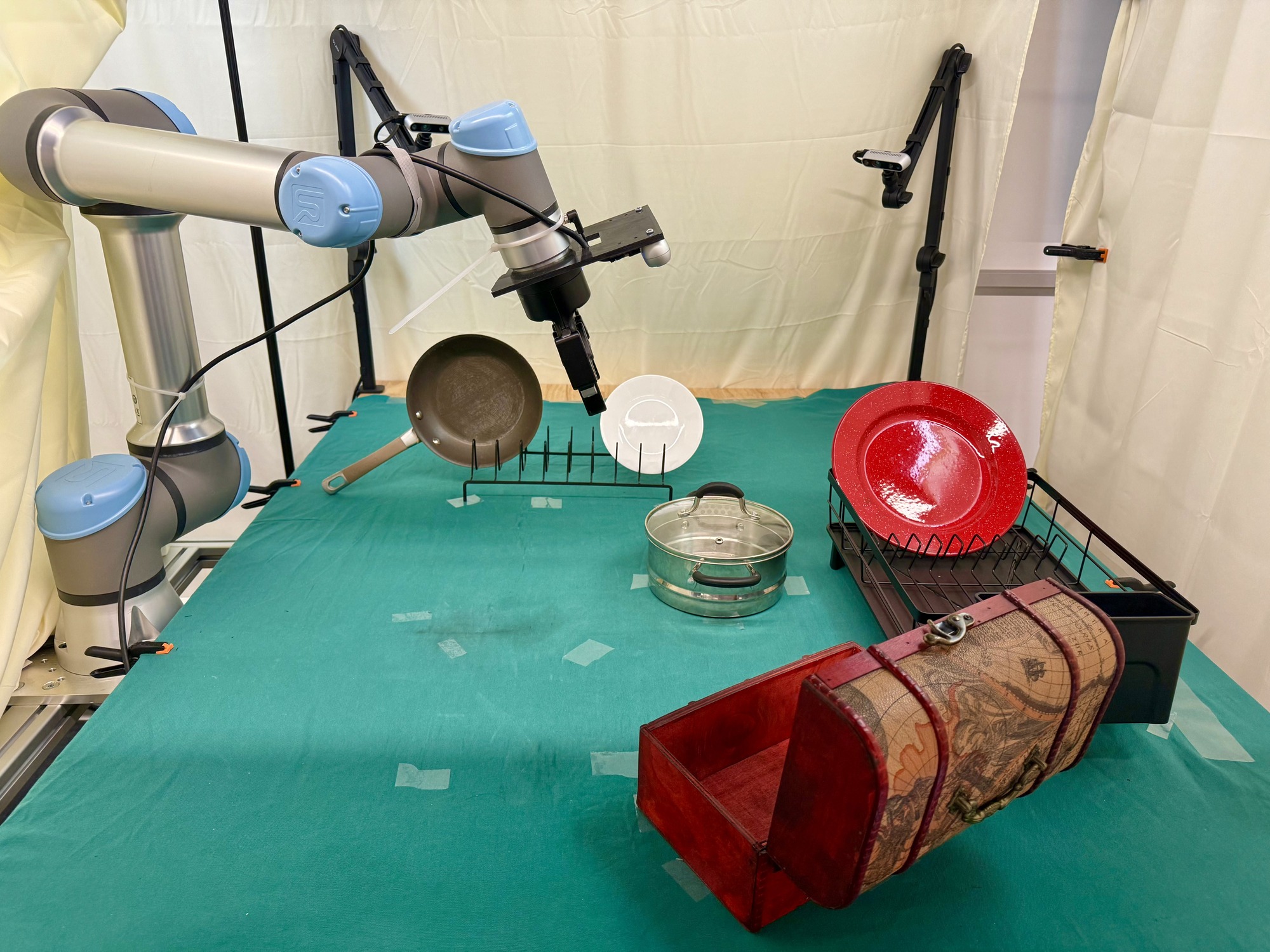}
    \includegraphics[width=.35\linewidth]{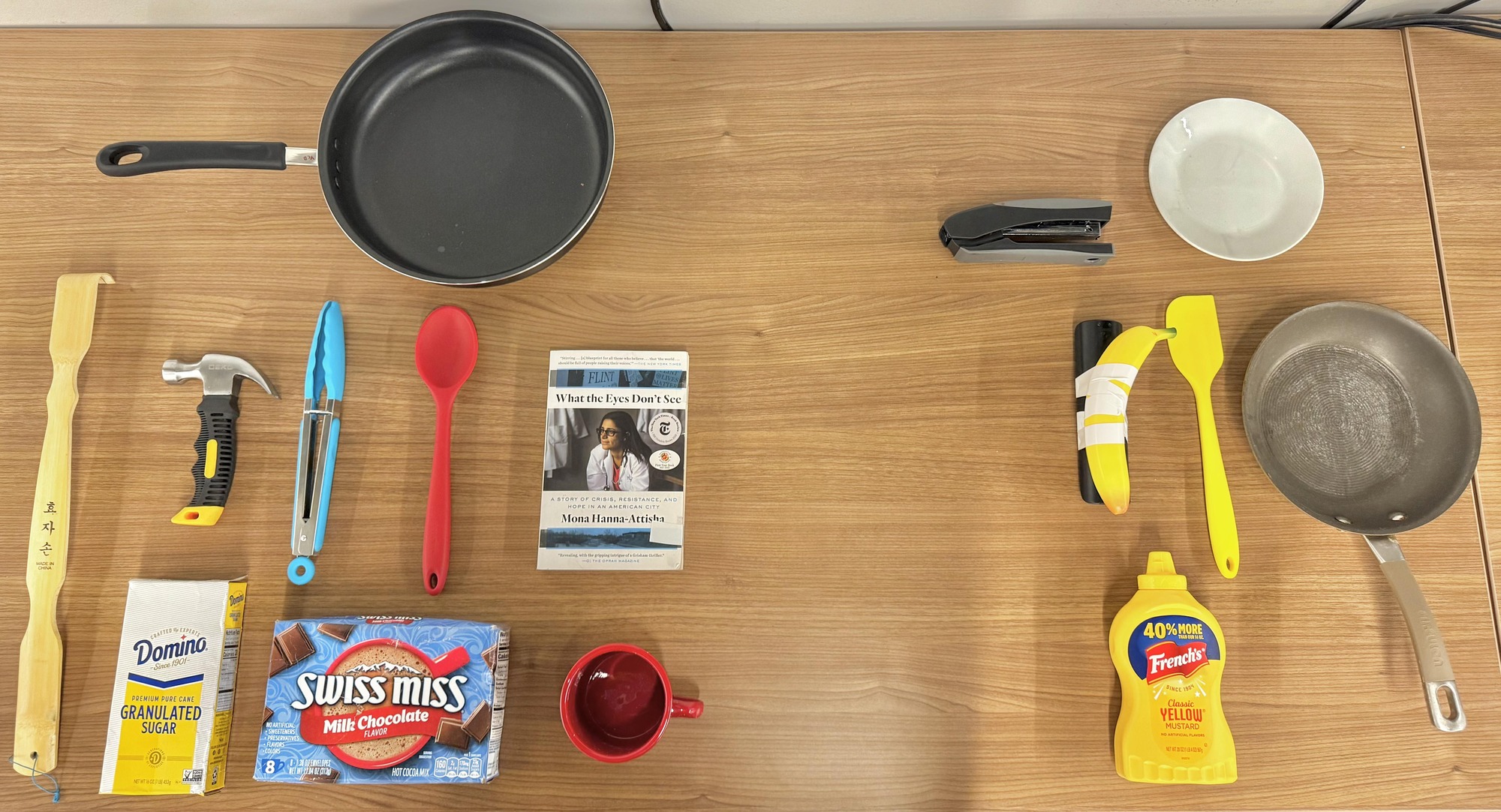}

    \caption{Our setup. We have a Realsense D455 camera, During our tests, we use a large number of background objects such as the two dish racks and the wooden chest to reflect a real world scenario. The right image shows all objects used in experiments. Left side are novel objects, right side are original objects.}
    \label{fig:robot_setup}
    \vspace{-0.1in}
\end{figure}

In our real-world experiments, we use an UR5e robot arm with a Robotiq-85 gripper. 
The arm is equipped with a Realsense D455 mounted to the front of the gripper (see Fig.~\ref{fig:robot_setup}). 
We run all experiments on a single Nvidia RTX A4000 GPU. We use the default Nerfacto model from Nerfstudio \cite{Tancik_2023} for training without any hyperparameter tuning aside from stopping the training at 4,000 timesteps for object NeRFs and 20,000 timesteps for background NeRFs.

For our behavior cloning model, we use BAKU~\cite{haldar2024bakuefficienttransformermultitask}. 
The model takes image inputs to predict the next action for the robot to take. 
We do not allow BAKU to access gripper position for any of the trials. 
For some tasks, we also added color jitter to combat auto-exposure and shadows. 

\begin{table*}[h]
    \caption{Task success rates for novel objects ($\uparrow$)}
    \label{table:success_rates}
    \renewcommand{\tabcolsep}{4pt}
    \renewcommand{\arraystretch}{1.2}
    \centering
    \tiny
    \begin{tabular}{@{}lccccccccc@{}}
    \toprule
    & \multicolumn{3}{c}{\textbf{Remove Object}} & \multicolumn{2}{c}{\textbf{Place in Chest}} & \multicolumn{1}{c}{\textbf{Serve on Plate}} & \multicolumn{1}{c}{\textbf{Cook with Pot}} & \multicolumn{2}{c}{\textbf{Lift up High}}\\
    \cmidrule(lr){2-4}
    \cmidrule(lr){5-6}
    \cmidrule(lr){7-7}
    \cmidrule(lr){8-8}
    \cmidrule(lr){9-10}
    \textbf{Methods} & \text{Blue Tongs} & \text{Red Spoon} & \text{Wood Stick} & \text{Swiss Miss Box} & \text{White Book} & \text{Large Pan} & \text{Sugar Box} & \text{Red Mug} & \text{Hammer} \\
    \midrule
    \textbf{Ours} & \textbf{70\%} & \textbf{100\%} & \textbf{80\%} & \textbf{90\%} & \textbf{60\%} & \textbf{100\%} & \textbf{80\%} & \textbf{90\%} & \textbf{100\%}\\
    \text{Default} & 0\% & 0\% &0\% & 10\% & 10\% & 0\% & 0\% & 0\% & 90\%\\
    \text{GenAug} & 10\% & 0\% & 0\% & 50\% & 0\% & 0\% & 0\% &  0\% & 0\% \\
    \text{Instruct Pix-to-Pix} & 40\% & 50\% & 10\% & 50\% & 0\% & \textbf{100\%} & 0\% & 0\% & 0\%\\
    \text{Roboagent} & 30\% & 20\% & 0\% & 80\% & \textbf{60\%} & 0\% & 0\% & 0\% & 10\%\\
    \text{Dreambooth} & 60\% & 0\% & 60\% & 20\% & 40\% & 0\% & 0\% & \textbf{90\%} & 0\%\\
    \bottomrule
    \end{tabular}
    \vspace{-0.2in}
\end{table*}
\subsection{Baselines}
While there has been substantial work with diffusion-based models in robot data augmentation~\cite{bharadhwaj2023roboagentgeneralizationefficiencyrobot,chen2023genaugretargetingbehaviorsunseen,mandi2023cactiframeworkscalablemultitask,Yu2023ScalingRL,Kapelyukh_2023}, to the best of our knowledge, we are only aware of one project that made their augmentation code available to the public: \textbf{GenAug} ~\cite{chen2023genaugretargetingbehaviorsunseen} which uses a diffusion inpainter to randomize textures and change objects. 
As such, we compare directly to GenAug's diffusion approach. 
We also compare to a common image-editing model \textbf{Instruct Pix-to-Pix}~\cite{brooks2023instructpix2pixlearningfollowimage}. In this baseline we prompt the model to edit the image by placing a novel object at the same position as the original object. We also wrote our own implementation of \textbf{Roboagent}~\cite{bharadhwaj2023roboagentgeneralizationefficiencyrobot}'s data augmentation method and compare to that. Finally, inspired by \cite{chen2024roviaugrobotviewpointaugmentation}, we finetune a Stable Diffusion Inpainter using Dreambooth \cite{ruiz2022dreambooth} to each of our novel objects with the same images used to train our NeRF models.
\subsection{Real World Tasks}
For the \textbf{Remove Object} task, our initial object is a yellow spatula that rests on top of a small frying pan. The robot grasps the spatula and lifts it from the pan then drops it to the left of the pan. For this task we test on 3 novel objects: blue tongs, a red spoon, and a long wooden stick. These objects show how well NeRF-Aug can generalize to different colors and lengths of objects. For this task, we used 30 demonstrations.\\

\indent For the \textbf{Place in Chest} task, our initial object is a mustard bottle which the robot grasps and places into a wooden chest. We tested on two novel objects: a Swiss Miss box and a paperback book. These objects differ in geometry and texture from the original object. We used 30 demonstrations.\\

\indent For the \textbf{Serve on Plate} task, our initial object is a small white plate that we place a pear onto. We then replace the white plate with a black pan about 2 times larger than the plate. This tests how well augmentation techniques can adapt to different sizes of objects. We used 23 demonstrations.\\

\indent For the \textbf{Cook with Pot} task, the robot removes the lid from a small pot and places it down on the table. It then picks up a stapler and places it in the pot. Our novel object is a sugar box that replaces the stapler. The sugar box has a more complicated texture than the stapler allowing us to evaluate how different textures affect policy performance as well as how well NeRF-Aug works in multi-step tasks. We use 10 demonstrations.\\

\indent For the \textbf{Lift up High} task, the robot grasps a banana and lifts it into the air. Our novel objects were a red mug and a small yellow-black hammer in place of the banana. The mug is much wider but not as long as the banana. The hammer is much thinner but heavier than the banana. In this case, we tested how different geometries and weights can affect policy performance. We use 10 demonstrations.

\begin{table*}[t] 
    \centering
    \tiny
    \caption{Time To Create New Data in Minutes Including NeRF Model Training ($\downarrow$)}
    \label{table:times1}
    \renewcommand{\tabcolsep}{2pt}
    \renewcommand{\arraystretch}{1.2}
    \begin{tabular}{@{}l c c c c c c c c c@{}}
    \toprule
    & \multicolumn{3}{c}{\parbox{2cm}{\centering \textbf{Remove Object}}} 
    & \multicolumn{2}{c}{\parbox{2cm}{\centering \textbf{Place in Chest}}} 
    & \multicolumn{1}{c}{\parbox{1cm}{\centering \textbf{Serve on Plate}}} 
    & \multicolumn{1}{c}{\parbox{2cm}{\centering \textbf{Cook with Pot}}} 
    & \multicolumn{2}{c}{\parbox{2cm}{\centering \textbf{Lift up High}}} \\
    \cmidrule(lr){2-4}
    \cmidrule(lr){5-6}
    \cmidrule(lr){7-7}
    \cmidrule(lr){8-8}
    \cmidrule(lr){9-10}
    \textbf{Methods} & \text{Blue Tongs} & \text{Red Spoon} & \text{Wood Stick} & \text{Swiss Miss Box} & \text{White Book} & \text{Large Pan} & \text{Sugar Box} & \text{Red Mug} & \text{Hammer} \\
    \midrule
    \textbf{Ours (w/o inpainter)} & 12.77 & 15.0 & 12.8 & 13.2 & 15.2 & 12.8 & 11.4 & 5.87 & 5.36\\
    \textbf{Ours (w/ inpainter)} & 30.9 & 33.1 & 31.0 & 44.5 & 46.5 & 44.7 & 38.7 & 24.9 & 24.4\\
    \text{GenAug} & 556 & 548 & 555 & 572 & 566 & 480 & 419 & 116 & 117 \\
    \text{Instruct Pix-to-Pix} & 73.5 & 73.6 & 70.4 & 75.4 & 76.5 & 64.7 & 55.9 & 15.7 & 15.8\\
    \text{Roboagent} & 1858 & 1835 & 1843 & 1913 & 1906 & 1633 & 1423 & 402 & 408 \\
    \text{Dreambooth} & 751 & 738 & 746 & 777 & 781 & 660 & 526 & 160 & 156 \\
    \bottomrule
    \end{tabular}
    \vspace{-0.2in}
\end{table*}

\subsection{Comparison to Baselines}
As seen in Table~\ref{table:success_rates}, our method consistently outperforms other methods on all five tasks. Over the next best method, we achieve an average of 43\% increase in the "Remove Objects" task, a 5\% increase in the "Place In Chest" task, an 80\% increase on the "Cook with Pot" task, and a 50\% increase on the "Lift up High" task. Our method achieved a 85.6\% success rate compared to the next best method (Dreambooth) with a 30\% success rate.

\textbf{Which methods synthesize object locations and sizes accurately?} GenAug, Instruct Pix-to-Pix, and RoboAgent often output images of an object in either the wrong orientation or position. As seen in fig. ~\ref{fig:results2}, we prompted the baselines to inpaint a "red mug," only for the output image to consist of a sideways view of a mug that is much larger than the actual mug. In contrast, our method was able to accurately synthesize the mug from a top-down view. Being able to render an object at unconventional viewpoints, such as when an object is very close to the camera or in a birds-eye-view, is a major reason why our method performs better than the baselines (such as being able to lift the mug 90\% more often than any other method). Dreambooth seems more effective at synthesizing the correct object viewpoint, but will fail when the original object is significantly different in shape from the old object (see fig. \ref{fig:results2} where Dreambooth fails to inpaint a swiss miss box and a book when the original object was a mustard bottle). A wide degree of viewpoints is important because images where the object is very close to the camera or from a birds-eye-view form the majority of an expert demonstration. Additionally, all of the baselines fail to inpaint the right size of the novel object.\\

\indent\textbf{Which method has higher image consistency?} Moreover, we observe that aside from Dreambooth, the baselines are highly inconsistent in the color and presence of distractor objects from frame to frame. The inconsistency can be detrimental to multi-camera systems where these methods may render a different looking object for each camera image at the same time-step. See Fig.~\ref{fig:results1} for examples where the diffusion models inconsistently adds distractor objects in one image and no distractor objects later on. Roboagent especially seems to add many distractors to the background. On the other hand, NeRF-Aug keeps the same colors and does not add distractor objects. \\

\indent\textbf{How do the methods compare in synthesizing texture and color?} We see that NeRF-Aug is able to accurately synthesize the exact color and texture of the novel object. Our most complicatedly textured objects were the Swiss Miss Box and Sugar Box, both of which NeRF-Aug was able to accurately synthesize down to the text on the box. In this area, GenAug fails by often incorporating the color of the original object into the modified image. Instruct Pix-to-Pix frequently makes the background of the frames the same color as the object we prompted for. Dreambooth seems to synthesize the texture correctly, but often fails to get the exact shade of the novel object.

\subsection{Data augmentation speed}
\begin{table}
    \centering
    \caption{Breakdown of Minutes Taken for Each Part of Our Method}
    \label{table:times2}
    \small 
    \begin{tabular}{@{}l >{\centering\arraybackslash}m{2.0cm} 
                    >{\centering\arraybackslash}m{2.3cm} 
                    >{\centering\arraybackslash}m{2.6cm} 
                    >{\centering\arraybackslash}m{2.3cm} 
                    >{\centering\arraybackslash}m{2.3cm} @{}}
        \toprule
        \textbf{Dataset} & \textbf{Train Novel Object NeRF} & \textbf{Segmentation} & \textbf{Background NeRF Training + Inpainting} & \textbf{Novel Object NeRF Rendering} \\
        \midrule
        Remove Object  & 3.40 & 4.75 & 13.38  & 9.72 \\
        Place In Chest & 5.08 & 5.50 & 25.83 & 10.07 \\
        Serve On Plate & 3.51 & 5.15 & 26.68  & 9.35 \\
        Lift Up High   & 5.07 & 1.45 & 17.61  & 2.07 \\
        Cook with Pot  & 3.03 & 4.47 & 22.92  & 8.36 \\
        \bottomrule
    \end{tabular}
    \vspace{-0.2in}
\end{table}

\indent We show that our method is significantly faster than other methods, as seen in Table~\ref{table:times1}. Our NeRF model is actually able to render at twice the fps as the original video because of our 128 by 128 resolution. Namely, our method is 63.6\% faster than the second fastest method (P2P). 

We also notice that the majority of the time taken was from the segmentation and background inpainter component (refer to Table~\ref{table:times2}). 
This module only needs to run once per task as opposed to once per novel object because it deals with erasing the original object in the training videos which is shared across all novel objects. For this reason, we also calculated the time without the training and running of the inpainter, resulting in NeRF-Aug running 4.99 times faster than P2P. In Table~\ref{table:times1}, we show the time taken including the segmentation, training the background NeRF model, and inpainting of the background NeRF (`w/ inpainter') and excluding this component (`w/o inpainter'). We measure all times on a single Nvidia RTX A4000 graphics card.




\section{Conclusion}
We introduce NeRF-Aug, a data augmentation framework for gripper-camera robotic systems. Our method leverages the photorealism of NeRFs to replace training objects in expert demonstrations with novel objects. This allows us to create synthetic training data for novel objects that is virtually indistinguishable from data that would otherwise require a human to demonstrate. Through extensive quantitative and qualitative experiments on 5 real-world tasks with 9 different objects, we show that policies trained using NeRF-Aug data are consistently more successful at tasks involving novel objects than the baselines, while only requiring a fraction of the time to augment. Future research could explore other novel-view synthesis methods such as Gaussian Splatting and Plenoxels to generate similar augmentation frameworks.

\bibliography{example}  

\end{document}